# Pose and Motion from Omnidirectional Optical Flow and a Digital Terrain Map

Ronen Lerner, Oleg Kupervasser and Ehud Rivlin
Department of Computer Science
Technion - Israel Institute of Technology
Haifa 32000, Israel
Email: ronenl@cs.technion.ac.il
oleg_kup@yahoo.com
ehudr@cs.technion.ac.il

*Abstract*— An algorithm for pose and motion estimation using corresponding features in omnidirectional images and a digital terrain map is proposed. In previous paper, such algorithm for regular camera was considered. Using a Digital Terrain (or Digital Elevation) Map (DTM/DEM) as a global reference enables recovering the absolute position and orientation of the camera. In order to do this, the DTM is used to formulate a constraint between corresponding features in two consecutive frames. In this paper, these constraints are extended to handle non-central projection, as is the case with many omnidirectional systems. The utilization of omnidirectional data is shown to improve the robustness and accuracy of the navigation algorithm. The feasibility of this algorithm is established through lab experimentation with two kinds of omnidirectional acquisition systems. The first one is polydioptric cameras while the second is catadioptric camera.

## I. INTRODUCTION

Vision-based navigation algorithms has been a major research issue during the past decades. Two common approaches for the navigation problem are: *landmarks* and *ego-motion integration*. In the landmarks approach several features are located on the image-plane and matched to their known 3D location. Using the 2D and 3D data the camera's pose can be derived. Few examples for such algorithms are [1], [2]. Once the landmarks were found, the pose derivation is simple and can achieve quite accurate estimates. The main difficulty is the detection of the features and their correct matching to the landmarks set.

In ego-motion integration approach the motion of the camera with respect to itself is estimated. The ego-motion can be derived from the optical-flow field, or from instruments such as accelerometers and gyroscopes. Once the ego-motion was obtained, one can integrate this motion to derive the camera's path. One of the factors that make this approach attractive is that no specific features need to be detected, unlike the previous approach. Several ego-motion estimation algorithms can be found in [3], [4], [5], [6]. The weakness of ego-motion integration comes from the fact that small errors are accumulated during the integration process. Hence, the estimated camera's path is drifted and the pose estimation accuracy decrease along time. If such approach is used it would be desirable to reduce the drift by activating, once in a while, an additional algorithm that estimates the pose directly. In [7] such navigation-system is being suggested. In that work, like in this work, the drift is being corrected using a Digital Terrain Map (DTM). The DTM is a discrete representation of the observed ground's topography. It contains the altitude over the sea level of the terrain for each geographical location. In [7] a segment of the ground was reconstructed using 'structure-from-motion' (SFM) algorithm and was matched to the DTM in order to derive the camera's pose. Using SFM algorithm, which does not make any use of the information obtained from the DTM but bases its estimate on the flow-field alone, positions their technique under the same critique that applies for SFM algorithms [8].

The algorithm presented in the previous work [9] does not require an intermediate explicit reconstruction of the 3D world. By combining the DTM information directly with the images information it is claimed that the algorithm is well-conditioned and generates accurate estimates for reasonable scenarios with reasonable error sources.

Recently, an increasing interest in omnidirectional vision for applications in robotics could be noted. Technically, omnidirectional vision, sometimes also called panoramic vision, can be achieved in various ways. Examples include camera with extreme wide angle lenses ("fish-eye"), cameras with hyperbolic or parabolic mirrors mounted in front of a standard lens (catadioptric imaging), sets of cameras mounted in a ring-like or sphere-like configuration (polydioptric imaging), or an ordinary camera that rotates around an axis and takes a sequence of images that covers a field of view of 360 degrees [10], [11], [12], [13], [14], [15], [16], [17], [18], [19].

Omnidirectional vision provides a very large field of view, which has some useful properties. For instance, it enables the tracking of objects which are placed in different directions in the surrounding scene. It is well established that such variety of features facilitates the obtainment of a robust and accurate estimate of the camera pose. On the other hand, vision algorithms have to account for the specific properties of the particular omnidirectional imaging sensor setup in use. This may comprise theoretical and methodological challenges, as is the case for catadioptric vision. Here, the extreme geometrical distortions of the images caused by the parabolic or hyperbolic

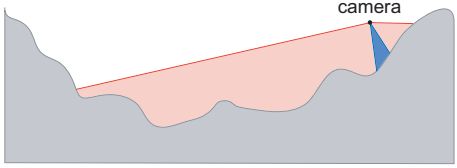

Fig. 1. When using an omnidirectional vision system a wide area of the terrain is visible (see the red area) even when the camera approaches a mountainside. When using a regular camera in similar scenario only small patch that is almost planar is observed (see the blue area).

mirror require a suitable adaptation of image interpretation methods.

The projection induced by an omnidirectional camera is the transformation from the 3D space to the image(s) plane. The least restrictive assumption that can be made about any camera model is that the inverse image of a point is a line in space. For many omnidirectional cameras, all such lines do not necessarily intersect in a single point. Their envelope is called a dia-caustic and represents a locus of viewpoints. If all the lines intersect in a single point, then the system has a single effective viewpoint and it is a central projection. In [20] a theorem is presented stating that a catadioptric camera has a single effective viewpoint if and only if the mirrors cross-section is a conic section. In any other case, including multiple cameras configurations, rotating camera systems and other shapes of mirrors, there is no single center of projection. The data acquired by such omnidirectional systems cannot be processed by vision algorithms that were developed under the single effective viewpoint assumption.

In this paper the navigation algorithm that was presented in [9] is extended to handle omnidirectional data. The most general case of non-central projection ("multi-optical center") is analyzed. The single center of projection case that was previously analyzed becomes a particular case of this general formulation when all optical centers are located in a single point. As was shown in [9], one of the most important factors that influence the robustness and the accuracy of the navigation algorithm is the complexity of the observed terrain. The extreme case, where only a planar segment of the terrain is visible, results in an ill-conditioned system which may lead to the failure of the algorithm. Whenever the navigating platform comes close to a mountainside in the terrain, such an ill-conditioned scenario might arise if a regular camera (not omnidirectional one) is used. However, when using an omnidirectional vision system, the rest of the terrain will still be visible even if the platform approaches one of the mountainsides (see Fig. 1). Therefore, more robust and accurate results can be achieved when using omnidirectional vision.

The paper continues as follows: Section II formally define the navigation problem. Section III derive the constraint for any corresponding features coming from two consecutive images along the trajectory. Experimental results are presented in section IV, and conclusions are drawn in section V.

## II. PROBLEM DEFINITION AND NOTATIONS

The problem can be briefly described as follows: At any given time instance $t$, a coordinates system $C(t)$ is fixed to an omnidirectional camera. At that time instance the camera is located at some geographical location $p(t)$ – a 3D vector, and has a given orientation $R(t)$ – an orthonormal rotation matrix, with respect to a global coordinates system $W$. $p(t)$ and $R(t)$ define the transformation from the camera's frame $C(t)$ to the world's frame $W$, where if $^C v$ and $^W v$ are vectors in $C(t)$ and $W$ respectively, then $^W v = R(t) {^C v} + p(t)$.

Considering two sequential time instances $t_1$ and $t_2$: the transformation from $C(t_1)$ to $C(t_2)$ is given by the translation vector $\Delta p(t_1, t_2)$ and the rotation matrix $\Delta R(t_1, t_2)$, such that $^{C(t_2)} v = \Delta R(t_1, t_2) {^{C(t_1)} v} + \Delta p(t_1, t_2)$. A rough estimates of the camera's pose at $t_1$ and of the ego-motion between the two time instances – $p_E(t_1)$, $R_E(t_1)$, $\Delta p_E(t_1, t_2)$ and $\Delta R_E(t_1, t_2)$ – are assumed to be known (the subscript letter "$E$" denotes that this is an estimated quantity). Such estimates can be obtained from dead-reckoning navigation system.

Also supplied is the optical-flow field. No special assumption is made on the omnidirectional acquisition system. It is assumed, however, that the system was fully calibrated. As a result, for each visible feature it is possible to compute its line of sight with respect to the camera system - $C$, which can be defined by a source point - $^C S_i$ and a unit-vector - $^C q_i$, oriented from the source point to the observed feature.

Using the above notations, the objective of the proposed algorithm is to estimate the true camera's pose and ego-motion: $p(t_1)$, $R(t_1)$, $\Delta p(t_1, t_2)$ and $\Delta R(t_1, t_2)$, using $n$ corresponding features from the optical-flow field $\{^C S_i(t_k), {^C q_i(t_k)}\}$ ($i=1\ldots n$, $k=1,2$), the DTM and the initial-guess: $p_E(t_1)$, $R_E(t_1)$, $\Delta p_E(t_1, t_2)$ and $\Delta R_E(t_1, t_2)$.

## III. THE NAVIGATION ALGORITHM

The following section describes a navigation algorithm which estimate the above mentioned parameters. The pose and ego-motion of the camera are derived using a DTM and the optical-flow field of two consecutive frames. Unlike the landmarks approach no specific features should be detected and matched. Only the correspondence between the two consecutive images should be found in order to derive the optical-flow field. As was mentioned in the previous section, a rough estimate of the required parameters is supplied as an input. Nevertheless, since the algorithm only use this input as an initial guess and re-calculate the pose and ego-motion directly, no integration of previous errors will take place and accuracy will be preserved.

The new approach is founded on the following observation. Since the DTM supplies information about the structure of the observed terrain, depth of observed features is being dictated by the camera's pose. Hence, given the pose and ego-motion of the camera, the optical-flow field can be uniquely determined. The objective of the algorithm will be finding the pose and ego-motion which lead to an optical-flow field as close as possible to the given flow field.

A single vector from the optical-flow field will be used to define a constraint for the camera's pose and ego-motion. Let $^WG \in \mathbb{R}^3$ be a location of a ground feature point in the 3D world. At two different time instances $t_1$ and $t_2$, this feature point is detected in the omnidirectional images and its lines of sight – $\{^CS(t_1), ^Cq(t_1)\}$ and $\{^CS(t_2), ^Cq(t_2)\}$ – are computed. Using an initial-guess of the pose of the camera at $t_1$, the line passing through $^CS(t_1)$ in direction of $^Cq(t_1)$ can be intersected with the DTM. Any ray-tracing style algorithm can be used for this purpose. The location of this intersection is denoted as $^WG_E$. The subscript letter "$E$" highlights the fact that this ground-point is the estimated location for the feature point, that in general will be different from the true ground-feature location $^WG$. The difference between the true and estimated locations is due to two main sources: the error in the initial guess for the pose and the errors in the determination of $^WG_E$ caused by DTM discretization and intrinsic errors. For a reasonable initial-guess and DTM-related errors, the two points $^WG_E$ and $^WG$ will be close enough so as to allow the linearization of the DTM around $^WG_E$. Denoting by $N$ the normal of the plane tangent to the DTM at the point $^WG_E$, one can write:

$$N^T(^WG - ^WG_E) \approx 0. \quad (1)$$

The true ground feature $^WG$ can be described using true pose parameters:

$$\begin{aligned} ^WG &= ^WS(t_1) + R(t_1) \cdot q(t_1) \cdot \lambda \\ &= R(t_1) \cdot (^CS(t_1) + q(t_1) \cdot \lambda) + p(t_1). \end{aligned} \quad (2)$$

Here, $\lambda$ denotes the distance between $^WS(t_1)$ and the feature point $^WG$. In the aforementioned equation we use the feature's transformed source point:

$$^WS(t_1) = R(t_1)^CS(t_1) + p(t_1). \quad (3)$$

Replacing (2) in (1) we get:

$$N^T[R(t_1) \cdot (^CS(t_1) + q(t_1) \cdot \lambda) + p(t_1) - ^WG_E] = 0. \quad (4)$$

From this expression, the distance of the true feature can be computed using the estimated feature location:

$$\lambda = \frac{N^{T\,W}G_E - N^{T\,W}S(t_1)}{N^TR(t_1)q(t_1)}. \quad (5)$$

In order to simplify notations, $R(t_i)$ will be replaced by $R_i$ and likewise for $p(t_i)$, $S(t_i)$ and $q(t_i)$ ($i=1,2$). $\Delta R(t_1, t_2)$ and $\Delta p(t_1, t_2)$ will be replaced by $R_{12}$ and $p_{12}$ respectively. The superscript describing the coordinate frame in which the vector is given will also be omitted, except for the cases were special attention needs to be drawn to the frames. Normally, $p_{12}$, $S_i$s and $q_i$s are in camera's frame while the rest of the vectors are given in the world's frame. Using the simplified notations, (5) can be assigned into (2) and after reorganization we get:

$$^WG = \frac{R_1q_1N^T}{N^TR_1q_1}{^WG_E} - \frac{R_1q_1N^T}{N^TR_1q_1}{^WS_1} + {^WS_1}. \quad (6)$$

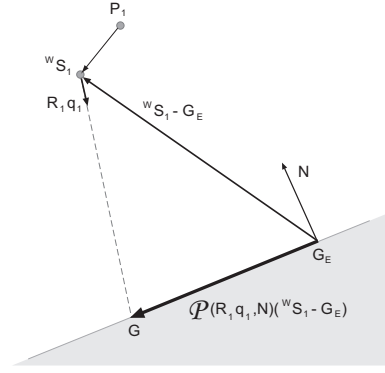

Fig. 2. Geometrical description of expression (9) using the projection operator (7)

In order to obtain simpler expressions, define the following projection operator:

$$\mathcal{P}(u, n) \doteq \left(\mathbf{I} - \frac{un^T}{n^Tu}\right) \quad (7)$$

This operator projects a vector onto the subspace normal to $n$, along the direction of $u$. As an illustration, it is easy to verify that $n^T \cdot \mathcal{P}(u, n)v \equiv 0$ and $\mathcal{P}(u,n)u \equiv 0$. By adding and subtracting $G_E$ to (6), and after reordering:

$$^WG = {^WG_E} + \left[\mathbf{I} - \frac{R_1q_1N^T}{N^TR_1q_1}\right]{^WS_1} - \left[\mathbf{I} - \frac{R_1q_1N^T}{N^TR_1q_1}\right]{^WG_E} \quad (8)$$

Using the projection operator, (8) becomes:

$$^WG = {^WG_E} + \mathcal{P}(R_1q_1, N)\left({^WS_1} - {^WG_E}\right) \quad (9)$$

The above expression has a clear geometric interpretation (see Fig.2). The vector from $G_E$ to $^WS_1$ is being projected onto the tangent plane. The projection is along the direction $R_1q_1$.

Our next step will be transferring $G$ from the global coordinates frame - $W$ into the first camera's frame $C_1$ and then to the second camera's frame $C_2$. Since $p_1$ and $R_1$ describe the transformation from $C_1$ into $W$, we will use the inverse transformation:

$$^{C_2}G = R_{12}R_1^T\left({^WG} - p_1\right) + p_{12}. \quad (10)$$

Assigning (9) into (10) gives:

$$^{C_2}G = R_{12} \cdot {^CS_1} + p_{12} + R_{12}\mathcal{L}\left({^WG_E} - {^WS_1}\right). \quad (11)$$

$\mathcal{L}$ in the above expression represents:

$$\mathcal{L} = \frac{q_1N^T}{N^TR_1q_1} \quad (12)$$

$q_2$ is a unit-vector pointing to the true ground-feature $G$. Thus, the vectors $q_2$ and $({^{C_2}G} - {^{C_2}S_2})$ should coincide. This observation can be expressed mathematically by projecting $({^{C_2}G} - {^{C_2}S_2})$ on the ray continuation of $q_2$:

$$^{C_2}G - {^{C_2}S_2} = q_2 \cdot \left(q_2^T \cdot ({^{C_2}G} - {^{C_2}S_2})\right) \quad (13)$$

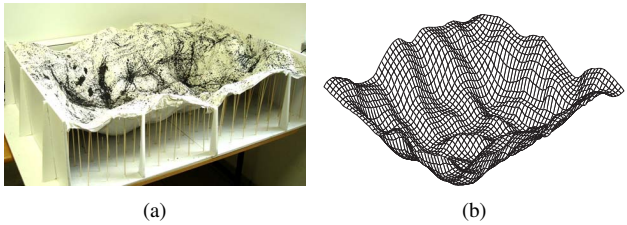

Fig. 3. (a) A 3D terrain model of horizontal dimension $115 \times 95$ cm. (b) The DTM was constructed by using a laser-based 3D-scanner. The spatial grid was 1mm (the one in the figure has a coarser grid for visualization purposes).

In expression (13), $q_2^T \cdot (^{C_2}G - {^{C_2}S_2})$ is the magnitude of $(^{C_2}G - {^{C_2}S_2})$'s projection on $q_2$. By reorganizing (13) and using the projection operator, we obtain:

$$\mathcal{P}(q_2, q_2) \cdot (^{C_2}G - {^{C_2}S_2}) = 0, \quad (14)$$

where:

$$\mathcal{P}(q_2, q_2) = \left[ I - q_2 \cdot q_2^T \right]. \quad (15)$$

$(^{C_2}G - {^{C_2}S_2})$ is being projected on the orthogonal complement of $q_2$. Since $(^{C_2}G - {^{C_2}S_2})$ and $q_2$ should coincide, this projection should yield the zero-vector. Plugging (11) into (14) yields our final constraint:

$$\mathcal{P}(q_2, q_2) \left[ R_{12} \cdot {^C S_1} + p_{12} + R_{12} \mathcal{L} \left( {^W G_E} - {^W S_1} \right) - {^{C_2} S_2} \right] = 0 \quad (16)$$

This constraint involves the position, orientation and the ego-motion defining the two frames of the camera. Although it involves 3D vectors, it is clear that its rank can not exceed two due to the usage of $\mathcal{P}$ which projects $\mathbb{R}^3$ on a two-dimensional subspace.

Such constraint can be established for each vector in the optical-flow field, until a non-singular system is obtained. Since twelve parameters need to be estimated (six for pose and six for the ego-motion), at least six optical-flow vectors are required for the system solution. Usually, more vectors will be used in order to define an over-determined system, which will lead to more robust solution. The reader attention is drawn to the fact that a non-linear constraint was obtained. Thus, an iterative scheme will be used in order to solve this system. For example, Newton-iterations which start from the rough estimate of the pose and motion parameters and iteratively converge to the least square solution can be performed. As was suggested in [21], M-estimator can be integrated into this scheme to increase its robustness in the presence of outliers.

## IV. EXPERIMENTAL RESULTS

Lab experimentation was performed using a real 3D model of a terrain and images from an omnidirectional acquisition system. The dimensions of the model were $115 \times 95$ cm with elevation variations as high as 32cm (see Fig.3(a)). A laser-based 3D-scanner was used to capture the terrain and build a DTM with a 1mm spatial grid (see Fig.3(b)).

Two types of omnidirectional acquisition systems were tested: a configuration of three regular cameras heading to different directions, and a catadioptric system with a parabolic mirror.

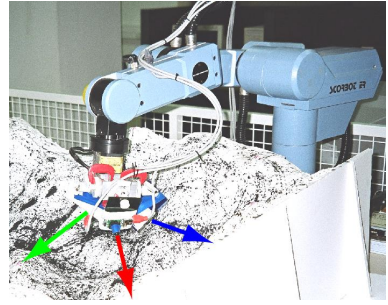

Fig. 4. Omnidirectional vision was obtained by a configuration of three cameras that were posed in different orientations.

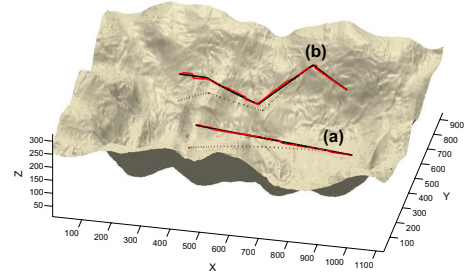

Fig. 5. Two of the tested trajectories. Trajectory *a* contains constant translational motion while trajectory *b* has significant changes in orientation. The true paths are marked by black solid line, while the pathes reconstructed by the algorithm are marked by red line. The black dotted lines represent the trajectories that would have been obtained in case the algorithm was not activated.

### A. Three Cameras Configuration

Three cameras with a wide field of view (80° each) were firmly attached to a robotic arm. Each camera was posed in a different orientation (see Fig. 4). Their internal parameters and relative pose parameters were accurately estimated as part of the system calibration phase. In each experiment the cameras configuration was moved along a different trajectory. The robotic arm allowed moving of the cameras in a controlled manner while also providing *true* measurements for the pose of the cameras at all time instances. Fig.5 shows examples of two of the trajectories evaluated. The first trajectory (*a* in the figure) contains constant translational motion with the orientation held constant. In the second trajectory (*b* in the figure) position and orientation of the cameras were changed significantly. Although highly accurate "ground-truth" data for the trajectory of the cameras was obtained from the robotic manipulator, this trajectory was corrupted using a simulated error model so that the "true" and the a priori trajectories drifted away with time. The error model drifted the trajectory position and orientation by 1 mm/sec and 0.7°/sec, respectively. In order to compensate for this drift, the proposed algorithm was called at 1 Hz rate. Whenever activated, this algorithm was supplied with the latest 3 images (one from each camera) and a previous image triplet that was captured 20mm away. The a priori information was derived from the available drifted pose at these two frames. Since 20mm baseline was desired, the algorithm was activated for the first time only after 3 seconds of movement. Later, it was periodically activated in 1 second gaps.

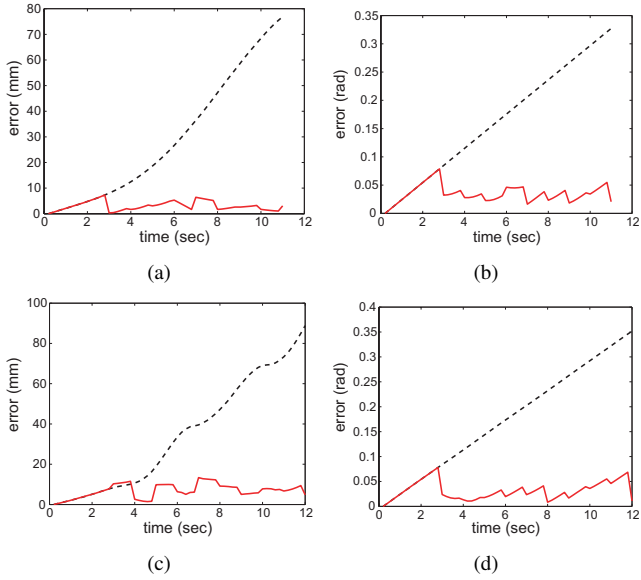

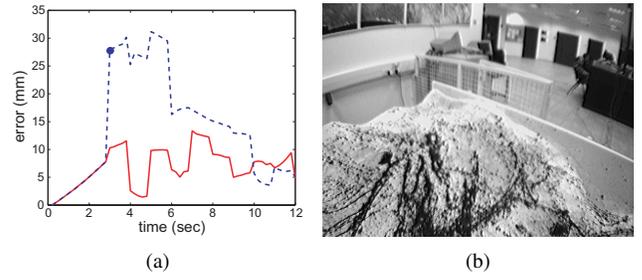

Fig. 7. (a) Translational errors of trajectory b when using 300 features coming from only one camera (blue dashed line) and when using 100 features from each of the three cameras (red solid line). (b) A frame captured by the single camera that was in use.

Fig. 6. Results for trajectories a (sub-figures (a) and (b)) and b (sub-figures (c) and (d)) when using the three cameras configuration. Position errors ((a) and (c)) and orientation errors ((b) and (d)) of the drifted path are marked with a black dashed line, and errors of the corrected path are marked with a red solid line.

During the experiments, gray-level images of $640 \times 480$ pixels were obtained from each of the three cameras. Correspondence between about 100 features per camera (300 features all together) was derived using the Lucas-Kanade tracking method [22], [23]. Features were not selected using an image-dependent algorithm, but rather, by using a regular grid spanned over the image-plane.

As shown in Figure 5, the algorithm converged to reasonable estimates for the navigation parameters along the two trajectories described above. The figure shows the "ground-truth" together with two trajectories computed using the error model: the first contains no updates while the second was updated periodically by using the proposed algorithm, at a 1 Hz rate. The figure clearly show that the corrected-path remains close to the true-path along the whole trajectory.

Figure 6 shows the position and orientation errors of the drifted and corrected paths for the two trajectories. It can be seen that the errors of the corrected path are kept small while the errors in the uncompensated path increase gradually. The saw-tooth shaped graph of the corrected path is characteristic: the orientation errors accumulate between updates but are strongly reduced each time the algorithm is applied.

In order to demonstrate the importance of the omnidirectional vision usage, the two trajectories were also reconstructed using 300 features coming from only one of the cameras, while the data from the other two cameras were ignored. Fig. 7(a) compares the translational accuracies that were obtained when using one vs. three cameras while reconstructing trajectory b. A clear advantage can be observed for the utilization of the omnidirectional configuration. In [9], the sensitivities of the proposed algorithm were studied. It was found that the obtained accuracy is highly related to the complexity of the observed terrain. The extreme case, where only a planar segment of the terrain is visible, results in an ill-conditioned system which leads to the failure of the algorithm. Whenever the navigating platform comes close to one of the mountainsides of the terrain, such an ill-conditioned scenario might happen if a regular camera (not omnidirectional one) is used. However, if using an omnidirectional vision system, then the rest of the terrain will still be visible even when approaching one of the mountainsides. Therefore, more robust and accurate results can be expected when using omnidirectional vision, as confirmed by Fig. 7(a). Note the blue dot in this figure. At that time instance, the algorithm performance was relatively poor for the single camera scenario since only small segment of the terrain was visible to that camera - Fig. 7(b).

### B. Catadioptric System

In the second experiment the three regular cameras were replaced by a single catadioptric system which is constructed of a parabolic mirror mounted in front of an orthographic camera (see Fig. 8(a)). Images of $1024 \times 768$ pixels were captured by this camera and 300 feature correspondences between two consecutive images were computed for the algorithm using the Lucas-Kanade method (see Fig. 8(b)). It should be noted that this tracking method is not optimal for catadioptric images due to the nature of the distortion of this kind of images. However, since the catadioptric system was first calibrated, these distortions can be computed and then cancelled. For each feature, a warped images can be rendered from the original images such that the local area of the feature appears as if it would be in a regular perspective camera. Next the Lucas-Kanade tracking method can be activated on these warped images with no special difficulty.

The translational and angular accuracies that were obtained during the two examined trajectories are presented in Figure 9. The slight deterioration in the algorithm performance (compared to its performance with the three cameras configuration) is probably due to the low resolution at the periphery of catadioptric images and due to the usage of the Lucas-Kanade tracking method directly on the distorted images.

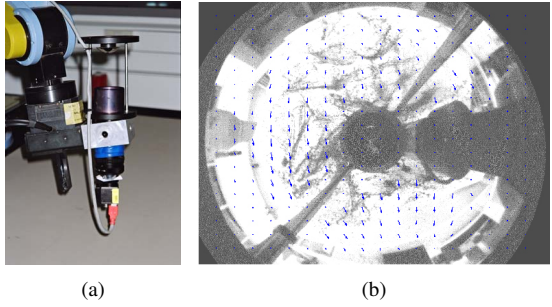

Fig. 8. (a) The catadioptric system that was used for omnidirectional vision in the second experiment. (b) An example for optical-flow field that was extracted for the algorithm. Each small blue arrow shows a corresponding couple.

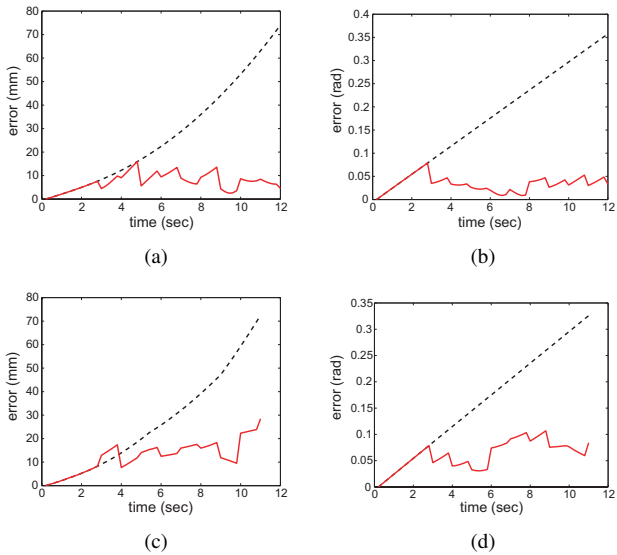

Fig. 9. Results for trajectories a (sub-figures (a) and (b)) and b (sub-figures (c) and (d)) when using the catadioptric system. Position errors ((a) and (c)) and orientation errors ((b) and (d)) of the drifted path are marked with a black dashed line, and errors of the corrected path are marked with a red solid line.

## V. CONCLUSIONS

An algorithm for pose and motion estimation using corresponding features in omnidirectional images and a DTM was presented. The DTM served as a global reference and its data was used for recovering the absolute position and orientation of the camera. The derived constraint eliminates the requirement for the commonly used assumption of single effective viewpoint. As a result, the presented algorithm is applicable for *all* omnidirectional acquisition systems. The performance of the presented algorithm was demonstrated using both polydioptric cameras and catadioptric camera. Both position and orientation estimates were found to be sufficiently accurate in order to bound the accumulated errors and to prevent trajectory drifts. Moreover, the utilization of omnidirectional data was shown to improve the robustness and accuracy of the navigation algorithm, compared to its counterpart algorithm for regular cameras. The improvement is attributed to the wide segment of the visible terrain. Such a segment tends to include much higher complexity than smaller segments which might be observed when using a regular camera.